\documentclass{article}

 \usepackage[dblblindworkshop,final]{neurips_2025}
 \workshoptitle{The First Workshop on Generative and Protective AI for Content Creation}
\usepackage{enumitem}
\usepackage[utf8]{inputenc} 
\usepackage[T1]{fontenc}    
\usepackage{hyperref}       
\usepackage{url}            
\usepackage{booktabs}       
\usepackage{amsfonts}       
\usepackage{nicefrac}       
\usepackage{microtype}      
\usepackage{xcolor}         
\usepackage{cite}
\usepackage{amsmath,amssymb,amsfonts}
\usepackage{algorithmic}
\usepackage{graphicx}
\usepackage{textcomp}
\usepackage{wrapfig}
\usepackage[most]{tcolorbox}
\usepackage{natbib}
\title{LegalWiz: A Multi-Agent Generation Framework for Contradiction Detection in Legal Documents}

\author{
    Ananya Mantravadi, Shivali Dalmia, Abhishek Mukherji, Nand Dave, Anudha Mittal \\
    Centific \\
    \And
    Olga Pospelova \\
    Amazon \\
    \texttt{\{ananya.mantravadi, shivali.dalmia, abhishek.mukherji,} \\
    \texttt{nandvinaykumar.dave, anudha.mittal\}@centific.com, posolga@amazon.com}
}
\begin{document}

\maketitle
\vspace{-18pt}
\begin{abstract}
Retrieval-Augmented Generation (RAG) integrates large language models (LLMs) with external sources, but unresolved contradictions in retrieved evidence often lead to hallucinations and legally unsound outputs. Benchmarks currently used for contradiction detection lack domain realism, cover only limited conflict types, and rarely extend beyond single-sentence pairs, making them unsuitable for legal applications. Controlled generation of documents with embedded contradictions is therefore essential: it enables systematic stress-testing of models, ensures coverage of diverse conflict categories, and provides a reliable basis for evaluating contradiction detection and resolution. We present a multi-agent contradiction-aware benchmark framework for the legal domain that generates synthetic legal-style documents, injects six structured contradiction types, and models both self- and pairwise inconsistencies. Automated contradiction mining is combined with human-in-the-loop validation to guarantee plausibility and fidelity. This benchmark offers one of the first structured resources for contradiction-aware evaluation in legal RAG pipelines, supporting more consistent, interpretable, and trustworthy systems.
\end{abstract}
\vspace{-8pt}
\section{Introduction}
\vspace{-8pt}
\label{sec:intro}

Large language models (LLMs) are increasingly used to draft legal documents, synthesize policies, and answer regulatory questions. These tasks differ significantly from general-purpose generation or factoid question answering. Legal text is dense, contextual, and often contains contradictions \footnote{In our case, contradictions may also correspond to policy violations or IP infringements.}, either introduced inadvertently, or arising from evolving regulations, overlapping jurisdictions, or competing organizational priorities. In high-stakes domains like law, compliance, and policy drafting, such contradictions are not just nuisances. They can lead to regulatory non-compliance, contractual disputes, and unclear ownership of intellectual property.

LLMs still struggle to reason over these conflicts or reliably identify them. Retrieval-augmented generation (RAG) pipelines attempt to ground outputs in external evidence, but unresolved conflicts in that evidence, such as superseded laws or inconsistent interpretations, often pass unchecked into the output. LLMs answering legal queries without retrieval grounding hallucinated between 69\% and 88\% of the time, fabricating statutes or misapplying case law \citep{dahl2024hallucinating}. Even legal-domain RAG systems hallucinate on over \emph{17\%} of benchmark queries \citep{magesh2025hallucination}, largely due to the retrieval of \textit{irrelevant or conflicting} sources. When contradictions in input evidence go unresolved, generation models often merge them, producing legally unsound and potentially risky outputs.

To benchmark and mitigate these failures, we need controlled generation of contradictions. Unlike natural contradictions that emerge as side-effects or errors, we focus on intentional contradiction injection, i.e., the ability to generate legal-style documents that contain subtle, structured contradictions by design. These contradictions may occur within a document (self-contradictions) or across documents (pairwise contradictions). This kind of controllability is essential for evaluating whether models can detect, resolve, or reason over contradictions when needed. It also opens the door to robust stress-testing of RAG pipelines and contradiction-aware generation systems.

Contradiction detection remains difficult for both humans and machines due to the subtle, context-dependent nature of many conflicts and the scarcity of domain-specific datasets. Even state-of-the-art models like GPT-4 \citep{gpt_4o} and LLaMAv2 \citep{touvron2023llama2openfoundation} perform only slightly better than chance \citep{li2023contradoc}. These challenges are compounded by the difficulty of defining ground truth in long-form legal texts. Our framework incorporates a human-in-the-loop setup for post-generation QA and for annotating contradiction pairs to support robust evaluation.

\vspace{-8pt}
\subsection{Challenges in Contradiction Benchmarking}
\vspace{-7pt}
We highlight four major challenges that make contradiction analysis retrieval-augmented generation (RAG) both necessary and difficult:
\vspace{-6pt}
\begin{enumerate}
\item \textbf{Unrealistic Legal Language:}  Current contradiction benchmarks often use overly plain or generic sentences, whereas real legal documents contain formal, complex sentence structures, and use intricate and nuanced language. 
\item \textbf{Limited Contradiction Types:} Most datasets focus on binary contradiction/entailment, overlooking complex types like temporal misalignment, reversals in obligations, or conflicts in legal authority - common in legal documents.
\item \textbf{Lack of Cross-Document Contradictions:} Most prior work focuses on single-document contradictions, while real conflicts often span multiple contracts, policies, or filings. 
\item \textbf{Manual Contradiction Detection is Time-Intensive:} Detecting contradictions across contracts or policies is slow and error-prone, often requiring line-by-line review.
\vspace{-8pt}


\end{enumerate}
\vspace{-8pt}

\subsection{Our Contribution}  
\vspace{-5pt}
To address these challenges, we introduce a novel multi-agent \textbf{contradiction-aware benchmark generation framework} that supports controllable generation of legal-style documents:  
\vspace{-5pt}
\begin{enumerate}
    \item \textbf{Realistic Legal Language:} We generate synthetic documents that mirror real legal tone, structure, and metadata. The framework supports customizable domains and subdomains, and quality is evaluated via perplexity, LLM assessment, and human validation.

    \item \textbf{Rich Contradiction Taxonomy:} We explicitly model six types of contradictions (Table \ref{tab:contradiction-taxonomy}) that goes beyond simple negation and captures the kinds of conflicts legal professionals routinely confront.  

    \item \textbf{Pairwise Contradictions:} We enable the controlled creation of documents with both \textit{self-contradictions} (contradictions present within the document) and \textit{pairwise-contradictions} (contradictions present across documents) reflective of legal drafting norms.

    \item \textbf{Contradiction-aware Retrieval Evaluation:} We provide an automated mining mechanism that detects contradictions using NLI models and LLM-based reasoning with confidence-weighted hybrid scoring. This is followed by human-in-the-loop validation through annotations conducted on the generated contradiction pairs. 

    \item \textbf{Human-in-the-Loop Supervision:} Human validation is embedded throughout the pipeline, from verifying contradiction realism and fluency, to resolving ambiguous contradiction pairs, ensuring that generated corpora maintain legal fidelity and that detection benchmarks reflect real-world judgment complexity.
\end{enumerate}  


To our knowledge, this is one of the first multi-agent generation framework for creating legal-style documents with controllable contradictions, designed to simulate the types of conflicts that arise in real-world legal, regulatory, and compliance workflows. These synthetic corpora allow us to evaluate how well systems detect and reason over contradictions, providing a foundation for more reliable document generation, retrieval filtering, and contradiction-aware evaluation in high-stakes domains.

\begin{table}[h]
\tiny
\begin{tabular}{|p{2cm}|p{5cm}|p{5cm}|}
\hline
\textbf{Contradiction Type} & \textbf{Description} & \textbf{Example} \\
\hline
\textbf{Temporal} & Contradicts date or time of an event. & \textit{“Starts Jan 15”} vs. \textit{“Starts end of Q1”} \\
\hline
\textbf{Numerical} & Conflicting numbers, values, or percentages. & \textit{“\$12M surplus”} vs. \textit{“\$5M deficit”} \\
\hline
\textbf{Authority} & Different source or issuer of a statement. & \textit{“Issued by Compliance Office”} vs. \textit{“Issued by Strategy Unit”} \\
\hline
\textbf{Process} & Conflicting procedures or operational routes. & \textit{“Submit via HR portal”} vs. \textit{“Submit through admins”} \\
\hline
\textbf{Policy Reversal} & One statement negates the other directly. & \textit{“Remote work mandatory”} vs. \textit{“Remote work not permitted”} \\
\hline
\textbf{Specificity} & One statement is more general or narrow than the other. & \textit{“Applies globally”} vs. \textit{“Applies only to APAC”} \\
\hline
\end{tabular}
\caption{Taxonomy of Contradiction Types}
\label{tab:contradiction-taxonomy}
\vspace{-8pt}
\end{table}
\vspace{-8pt}
\section{Background and Related Work}
\vspace{-5pt}
\subsection{Contradictions in Generation and Retrieval}

Natural Language Inference (NLI) involves determining the logical relationship between two sentences as entailment, neutrality, or contradiction. Contradiction detection is a focused subtask that identifies when two statements cannot simultaneously be true under the same interpretation. Contradictions can occur either \textit{within} a single document (self-contradictions) or \textit{across} multiple documents on related content (pairwise contradictions). In retrieval-augmented generation (RAG), unresolved conflicts like outdated statutes, overlapping jurisdictions, or inconsistent interpretations can lead models to produce factually incorrect or logically inconsistent outputs. Left unfiltered, these contradictions represent a high-risk form of hallucination in legal and policy settings.


Despite growing interest, existing contradiction benchmarks are poorly suited for stress-testing generative systems. Early work in NLI introduced datasets like SNLI \citep{bowman2015large}, MNLI \citep{williams2017broad}, and ANLI \citep{williams2020anlizing}, but these focus on isolated sentence pairs and lack the linguistic and structural complexity of real documents. More domain-specific resources such as ContractNLI \citep{koreeda2021contractnli} are limited to clause-level reasoning. ContraDoc \citep{li2023contradoc} and ECON \citep{jiayang2024econ} move toward document- and retrieval-scale evaluation. ContraDoc showed that even GPT-4 struggles with subtle internal inconsistencies in long documents, and that trained annotators also miss many contradictions. ECON distinguishes factoid and answer conflicts, showing that models often default to internal priors when evidence is inconsistent. \citet{gokul2025contradiction} extend this work to synthetic RAG pipelines by generating controlled contradictions and testing LLMs as context validators, but consider only three types: self, pairwise, and conditional.

\begin{figure*}[h!]
    \centering 
    \includegraphics[width=1\textwidth]{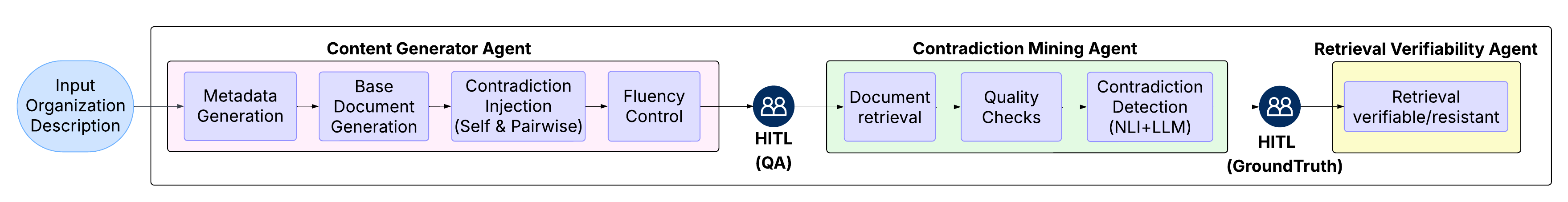} 
    \caption{LegalWiz - Workflow} 
    \label{fig:your_figure_label} 
\vspace{-10pt}
\end{figure*}

\subsection{Multi-Agent and Generative Systems}
Recent literature has explored content generation through multi-agent architectures. \citet{musumeci2024llm} propose a multi-agent architecture that automates the drafting of semi-structured public administration documents by decomposing templates, retrieving relevant information, and generating text iteratively to reduce hallucinations, demonstrating how agent specialization can yield more structured and reliable outputs. \citet{xie2024large} survey large multimodal agents and emphasize perception, planning, and memory as essential components for real-world deployment. \citet{wang2025talkstructurallyacthierarchically} introduce TalkHier, a collaborative LLM-MA framework that improves task performance through structured communication and hierarchical refinement. 

Current benchmarks leave open a critical gap in building realistic, controllable corpora that simulate the retrieval conflicts common in legal and compliance settings. Our agentic framework allows us to decouple generation, contradiction injection, and evaluation into specialized roles, making it easier to control the types of contradictions introduced and automate quality checks for the creation of benchmark-ready corpora.


\vspace{-8pt}
\section{System Architecture}
\vspace{-8pt}
\textbf{LegalWiz} is a modular multi-agent system designed for generating and evaluating legal documents enriched with controlled contradictions. It supports document creation, contradiction injection, fluency validation, hybrid contradiction mining, and retrieval-based fact-checking. All agents communicate via \texttt{Pyro4}-based remote procedure calls and are coordinated by a central orchestrator. The overall architecture is shown in Figure \ref{fig:your_figure_label}.


\textbf{LegalWiz} integrates three coordinated agents spanning document generation to verification. The \textbf{Content Generator Agent} creates structured metadata and base legal documents, injects self- and pairwise contradictions via instruction-tuned prompts, and validates fluency using relative perplexity while logging conflict metadata. The \textbf{Contradiction Mining Agent} combines an NLI model with an LLM judge, using semantic similarity filtering to detect intra- and inter-document conflicts. A \textbf{human-in-the-loop} review refines low-confidence cases. The \textbf{Retrieval Verifiability Agent} then assesses whether conflicting statements are externally verifiable, distinguishing retrieval-verifiable from retrieval-resistant contradictions.

\begin{figure*}[h!]
    \centering 
    \includegraphics[width=0.85\textwidth]{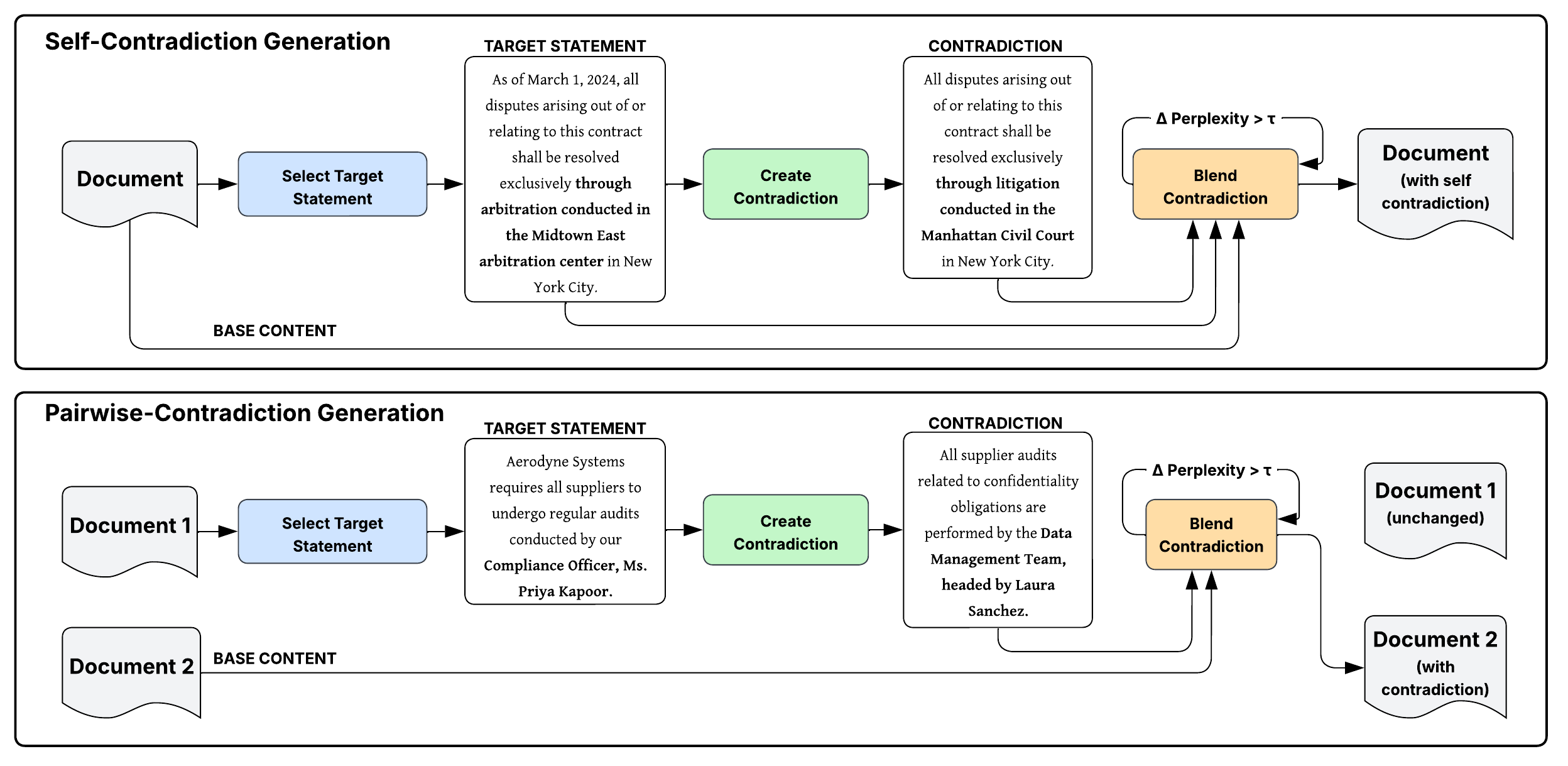} 
    \caption{Contradiction Generation} 
    \label{fig:contra} 
\end{figure*}
\vspace{-14pt}
\subsection{\textbf{Contradiction-Aware Content Generation Agent}}

\subsubsection{Metadata-Driven Base Document Generation}
To initiate generation, we construct a realistic organizational profile that serves context for generating metadata. The scope of legal domain is organized into five categories. Given the organization description and subdomain structure, metadata is generated containing \textit{title}, \textit{topic}, \textit{date}, \textit{department}, \textit{location}, \textit{document} \textit{type}, and \textit{authority level}. This structured metadata serves as the source for multi-paragraph legal documents generated in assertive, policy-oriented language (see Prompt: Base Document Generation in \ref{appendix:contentgen}). We discuss the exact organization profile and domain-subdomain structure using in our framework in section \ref{data}.
\vspace{-5pt}
\subsubsection{Perplexity-Based Fluency Control}
Perplexity is a fundamental metric in Natural Language Processing used to assess the quality of language models and, by extension, to gauge the fluency and coherence of generated text. A language model that can accurately predict the next words in a sequence is more likely to generate fluent and grammatically correct text. 
Given a document represented as a sequence of tokens $w_1, w_2, \ldots, w_N$, the perplexity is:  
\[
PP(w_{1:N}) = \exp\left(-\frac{1}{N}\sum_{i=1}^{N}\log P(w_i \mid w_{<i})\right).
\]  where $P(w_i \mid w_{<i})$ is the probability assigned to the $i$-th token given its preceding context $w_{<i}$, and $N$ is the total number of tokens in the document. A lower perplexity indicates that the model finds the text more predictable (i.e., fluent) under its learned distribution. To ensure fluency preservation after contradiction injection, the perplexity of both the base and contradicted document is computed using pretrained language model GPT2 \citep{radford2019language}. Let $PPL_{\text{base}}$ denote the perplexity of the base document and $PPL_{\text{contr}}$ the perplexity after contradiction injection. If the perplexity $PPL > 22$, the generated document is rejected and re-tested. The relative change is defined as:
\[
\Delta_{\text{rel}} = \frac{PPL_{\text{contr}} - PPL_{\text{base}}}{PPL_{\text{base}}}
\]
\vspace{-10pt}
\subsubsection{Contradiction Injection}
To embed a \textit{self-contradiction} (see Figure \ref{fig:contra}) , a target sentence is first selected using an LLM prompt that identifies a factual, specific, and authoritative statement. A second few-shot prompt generates a direct and confident contradiction statement, explicitly avoiding hedging language (e.g., “however,” “may,” “sometimes”). Finally, a third prompt blends the contradiction seamlessly into the document, preserving coherence, tone, and style. The prompts used are mentioned in the appendix \ref{appendix:contra}).

\textit{Pairwise contradictions} are created by selecting a target statement from a source document $d_1$ and generating a corresponding contradictory statement as described earlier. The contradiction is embedded into a separate document $d_2$, ensuring stylistic and structural consistency while omitting the original target statement. This results in a document pair $\langle d_1, d_2 \rangle$, where $d_2$ contradicts a key assertion in $d_1$. The defined thresholds are validated, and any violation triggers document regeneration. This ensures fluent integration of contradictions while preventing unnatural degradation from language model nondeterminism. 
\vspace{-0.1cm}
\begin{equation}
\Delta_{\text{rel}}^{\text{self}} \leq 0.05 \quad \text{(self-contradiction, $\leq$ 5.5\% increase)}
\end{equation}
\begin{equation}
\Delta_{\text{rel}}^{\text{pair}} \leq 0.075 \quad \text{(pair contradiction, $\leq$ 7.5\% increase)}
\end{equation}
\begin{equation}
PPL_{\text{contr}} \leq 22.0 \quad \text{(absolute cap for sanity)}
\end{equation}

We conduct a human evaluation of generated documents containing injected contradictions using a 5-point Likert scale. The results show strong average ratings for fluency (4.2), specificity (4.7), and coherence (4.5), indicating the documents were well-written, detailed, and structurally consistent. Legitimacy scored slightly lower (3.5), suggesting that some documents lacked the authority expected of authentic legal texts. Contradictions were detected in only 43\% of the cases, and when identified, were rated as subtle and realistic. Although our injected contradictions are naturalistic, they remain challenging for even human reviewers to reliably detect.
\vspace{-5pt}
\subsection{\textbf{Contradiction Mining Agent}}
Long-form legal documents often contain subtle inconsistencies which makes contradiction identification challenging for both humans and models. The Contradiction Mining Agent identifies high-likelihood contradiction pairs, which are then validated by human annotators. This agent ensures: 1) benchmark quality by verifying non-trivial, detectable contradictions and 2) provides a baseline for assessing RAG systems’ ability to recognize conflicts before generation.

NLI models are fast and widely used but they often struggle with spurious correlations, brittleness, and the difficulty of capturing nuanced reasoning. LLMs, on the other hand, offer richer reasoning but are computationally expensive. To identify contradictions at scale across synthetic legal documents, we combine fast retrieval-style filtering with multi-stage entailment reasoning. Our approach avoids exhaustive $\mathcal{O}(n^2)$ sentence pair comparison.

\begin{enumerate}
\item \textbf{Top-$k$ Semantic Filtering:} For each source sentence, cosine similarity is computed against candidate targets using \texttt{msmarco-distilbert-base-v3} \citep{reimers-gurevych-2019-sentence}, embeddings, fine-tuned for passage retrieval. The top-$k$ (k=5) pairs above threshold $\theta_s$ are retained, while short sentences (<10 words), numbers, and bullet points are filtered out. Duplicate pairs are removed via text-based hashing, effectively reducing the search space while preserving high-relevance candidates.

\item \textbf{NLI-Based Contradiction Classification:}
The filtered sentence pairs are evaluated using a pretrained Natural Language Inference (NLI) model (\texttt{facebook/bart-large-mnli}) \citep{yin2019benchmarkingzeroshottextclassification} to assign a label $\ell_{\text{NLI}} \in {\text{contradiction}, \text{neutral}, \text{entailment}}$ with confidence $p_{\text{NLI}} \in [0,1]$. Pairs labeled as contradictions or with $p_{\text{NLI}} \leq \theta_{\text{conf}}$ ($\theta_{\text{conf}}=0.7$) are forwarded for further assessment by a large language model (LLM).

\item \textbf{LLM-Based Contradiction Judgment:}
For the retained pairs, GPT-4o \citep{gpt_4o} is prompted (see appendix \ref{appendix:judge}) to assess whether both statements can be simultaneously true. It returns a binary contradiction label $\ell_{\text{LLM}} \in {0,1}$, a justification, and a confidence score $p_{\text{LLM}} \in [0,1]$ reflecting its certainty.

\item \textbf{Confidence-Weighted Hybrid Scoring:} A hybrid contradiction score is computed as:
\[s_{\text{hybrid}} = w_{\text{NLI}} \cdot \ell_{\text{NLI}} + w_{\text{LLM}} \cdot \ell_{\text{LLM}}, \]
where $\ell_{\text{NLI}}, \ell_{\text{LLM}} \in \{0,1\}$ represent binary contradiction labels (1-contradiction), and weights $w_{\text{NLI}}, w_{\text{LLM}}$ are derived from the model confidences:
\[w_{\text{NLI}} = \frac{p_{\text{NLI}}}{p_{\text{NLI}} + p_{\text{LLM}}}, \quad
w_{\text{LLM}} = \frac{p_{\text{LLM}}}{p_{\text{NLI}} + p_{\text{LLM}}}\]
ensuring $w_{\text{NLI}} + w_{\text{LLM}} = 1$. A pair is classified as contradictory if $s_{\text{hybrid}} > \tau$, where $\tau$ is a decision threshold ($\tau = 0.5$).

\end{enumerate}

\textbf{Human-in-the-Loop (HITL)} annotations are vital for ensuring both the realism of contradiction generation and the reliability of contradiction detection in long-form legal texts. Automated methods can flag inconsistencies, but human expertise is required to validate plausibility and establish reliable ground truth. To quantify annotation reliability, we measured inter-annotator agreement (IAA) across self- and pairwise contradictions: for self, agreement was \(96.0\%\) with Cohen’s \(\kappa=0.9143\), Krippendorff’s \(\alpha=0.9146\); for pairwise, \(94.77\%\) with \(\kappa=0.828\), \(\alpha=0.8278\). Samples below a \(90\%\) threshold are adjudicated by a subject matter expert. Embedding such human judgment ensures dataset fidelity, model robustness, and legally sound contradiction detection.

\par This staged framework enables interpretable contradiction detection in semi-structured documents, supporting both self- and pairwise analyses. Serving as a mining assistant rather than an oracle, it filters candidate pairs to a high-confidence subset for human verification, ensuring contradictions are both machine-detectable and human-confirmable while balancing scale and quality.
\vspace{-5pt}

\subsection{\textbf{Retrieval Verifiability Agent}}  
A key challenge in evaluating contradiction resolution is distinguishing failure sources: retrieval gaps (missing evidence) versus reasoning gaps (requiring inference beyond retrieval). Existing benchmarks blur these distinctions, identification of RAG pipeline breakdowns. This agent addresses this gap by determining each contradiction pair $(s_1, s_2)$ is \emph{retrieval-verifiable} or \emph{retrieval-resistant}:  
\vspace{-7pt}
\begin{itemize}  
\item \textbf{Retrieval-verifiable}: Contradictions that can be resolved using reliable, publicly available evidence that  \textit{(should be)}retrievable  by a RAG system.  
\item \textbf{Retrieval-resistant}: Contradictions lacking sufficient supporting evidence, requiring reasoning or human judgment beyond retrieval.
\vspace{-5pt}

\end{itemize}  

This distinction makes contradiction evaluation \textit{diagnostic} rather than descriptive. For example, consider two clauses: ``Termination requires 30 days’ notice’’ vs. ``Termination requires 90 days’ notice.’’ If a statute mandates 90 days, the contradiction is \textbf{retrieval-verifiable}, indicating a retrieval weakness. If no statute exists, it is \textbf{retrieval-resistant}, reflecting limitations in retrieval and the need for deeper reasoning. Labeling contradictions this way localizes errors, making evaluation actionable for improving legal RAG systems.

\vspace{-8pt}
\section{Dataset Construction and Characteristics} \label{data}
\vspace{-10pt}
To evaluate contradiction detection in a controlled yet realistic setting, a golden standard of 50 synthetic legal documents is constructed around a fictional multinational company, Aerodyne Systems, which designs aerospace technologies and operates globally under strict confidentiality agreements. The corpus spans five legal domains—contract law, compliance and regulation, internal policy and governance, dispute resolution and litigation, and terms and service management—ensuring domain realism and structured variation across organizational functions.

\textbf{Organization description -} \textit{“A fictional multinational aerospace company called Aerodyne Systems that designs and manufactures proprietary aerospace technologies and operates across offices in the U.S., Europe, and Asia. It routinely engages in partnerships with external vendors, research institutes, and government agencies, requiring strict non-disclosure and licensing agreements.”}

After passing fluency checks, contradictions are systematically injected in a domain-specific manner. Each document in Dispute Resolution and Litigation and Terms and Service Management contains one self-contradiction, while an interleaving strategy within Contract Law, Internal Policy and Governance, and Compliance and Regulation modifies every second document to contradict the preceding one. This setup creates both intra- and inter-document conflicts, simulating realistic organizational scenarios. As shown in Table~\ref{tab:contradiction_distribution}, specificity and policy reversals dominate across contradiction types, with temporal inconsistencies prevalent in self-contradictions and specificity conflicts more common in pairwise cases, reflecting discrepancies in timelines, goals, and resource allocations.

\begin{table}[htbp]
\vspace{-9.5pt}
\centering
\begin{tabular}{lcc}
\hline
\textbf{Contradiction Type} & \textbf{Self} & \textbf{Pairwise } \\
\hline
Temporal & 12 & 18 \\
Numerical & 5 & 1 \\
Specificity & 6 & 14 \\
Policy Reversal & 17 & 9 \\
Authority & 2 & 5 \\
Process & 6 & 9 \\
\hline
Total & 48 & 56\\
\hline
\end{tabular}
\caption{Distribution of contradiction types for self and pairwise contradictions.}
\vspace{-15pt}
\label{tab:contradiction_distribution}
\end{table}

While contradictions were injected by category, unintended conflicts (e.g., temporal inconsistencies) occasionally arose. Human annotations tracked whether injected contradictions were successfully blended, created detectable conflicts, and were correctly identified, enabling precise evaluation of system robustness and sensitivity to spurious cases.

\vspace{-9pt}

\section{Results and Discussion}
\vspace{-9pt}
We benchmarked three detectors: \textit{NLI}, \textit{LLM Judge}, and \textit{Hybrid (NLI+LLM)} on both self-contradictions and pairwise contradictions. As discussed in Section~3.2, \textit{NLI} lacks contextual depth, while \textit{LLM Judge} suffers from prompt sensitivity, limited evidence verification, hallucination, and overconfidence. The \textit{Hybrid} approach integrates NLI predictions with LLM judgments using a confidence-weighted scoring strategy to produce a consolidated contradiction label. This combination was adopted to balance NLI’s scalability and precision with the LLM’s richer contextual reasoning capabilities, leading to more reliable contradiction detection. Contradictions were mined using three detectors, each identifying partially overlapping sets of pairs. Since no single detector achieved full coverage, a unified evaluation dataset was constructed:
\begin{itemize}
    \item Each candidate pair was uniquely defined by (\texttt{doc1\_chunk}, \texttt{doc2\_chunk}), normalizing whitespace and formatting for consistency. The union of all detector outputs ensured inclusion of every potential contradiction.
    \item Ground-truth injected contradictions were manually added to the union to assess detector recovery of known cases.
    \item Human annotators reviewed surrounding context and assigned binary labels (\texttt{human\_label}) indicating the presence or absence of contradiction.
\end{itemize}
\vspace{-8pt}
This process resolved inconsistencies, consolidated duplicates, and produced a unified gold-standard dataset comprising \textbf{100 self-contradiction pairs} and \textbf{306 pairwise-contradiction pairs}, with representative examples shown in Tables~\ref{tab:self_examples} and \ref{tab:pair_examples}.

\begin{table*}[h!]
\centering
\resizebox{\textwidth}{!}{
\tiny
\begin{tabular}{|p{5cm}|p{5cm}|c|c|}
\hline
\textbf{Target Statement} & \textbf{Contradiction} & \textbf{Hybrid Pred} & \textbf{Human Label} \\
\hline
Effective from March 1, 2024, any disclosure of sensitive information without proper authorization will be considered a breach of this agreement. &
Unauthorized disclosure of sensitive information will not be considered a breach of this agreement until May 15, 2024. &
TRUE & TRUE \\
\hline

Upon submission, summaries will automatically be distributed to the opposing party's legal team and the appointed mediator, ensuring no delays in mediation proceedings. &
Summaries must be reviewed and approved by both the Lead Counsel and Senior Mediation Advisor, Priya Deshmukh, before submission. &
TRUE & FALSE \\
\hline
It is critical to note that all joint development projects initiated after April 10, 2024, will adhere strictly to this framework. &
All joint development projects initiated after April 10, 2024, are exempt from adhering to this framework and will operate under an adaptive guideline model. &
FALSE & TRUE \\
\hline

Seconded employees will retain full ownership of any intellectual property developed during their tenure with the joint venture. &
This agreement ensures that all creative outputs remain the exclusive property of the individual employee, irrespective of their parent company obligations. &
FALSE & FALSE \\
\hline
\end{tabular}}
\caption{Examples of self-contradictions with hybrid predictions vs. human labels.}
\vspace{-10pt}
\label{tab:self_examples}
\end{table*}

\begin{table*}[h!]
\tiny
\centering
\resizebox{\textwidth}{!}{
\begin{tabular}{|p{5cm}|p{5cm}|c|c|}
\hline
\textbf{Doc1 Target Statement} & \textbf{Doc2 Contradiction} & \textbf{Hybrid Pred} & \textbf{Human Label} \\
\hline
Aerodyne Systems is committed to maintaining full transparency and compliance, submitting all required reports by March 30, 2024, without exceptions. &
Aerodyne Systems will submit the required regulatory compliance reports by April 15, 2024, ensuring alignment with updated industry standards. This submission deadline accommodates necessary internal auditing processes. &
TRUE & TRUE \\
\hline
It is imperative that all Aerodyne Systems employees and vendors understand that any breach of this agreement will result in immediate legal action and potential termination of contracts, as the Company maintains a zero-tolerance policy for violations. &
The vendor is authorized to disclose Confidential Information to third parties immediately upon the termination of this Agreement, without requiring prior written consent from Aerodyne Systems. &
TRUE & FALSE \\
\hline
Upon termination, the Vendor agrees to return or destroy all documents containing or reflecting Confidential Information and to confirm completion of this process in writing. &
The vendor is authorized to disclose Confidential Information to third parties immediately upon the termination of this Agreement, without requiring prior written consent from Aerodyne Systems. &
FALSE & TRUE \\
\hline
The Crisis Management Policy is interpreted and enforced by the Internal Policy and Governance team led by Fiona Zhang, Director of Internal Policy. &
The Internal Policy and Governance team, led by Dr. Ravi Kumar, Policy Compliance Officer, and supported by Mei Lin Tan, Legal Affairs Specialist, is responsible for overseeing the Conflict of Interest Declaration process. &
FALSE & FALSE \\
\hline
\end{tabular}}
\caption{Examples of pairwise contradictions with hybrid predictions vs. human labels.}
\label{tab:pair_examples}
\vspace{-14pt}
\end{table*}

The performance comparison of contradiction miners, summarized in Table 5, highlights clear differences across the three approaches. For self-contradiction detection, the NLI-only model achieves high recall (81.6\%) but suffers from low precision (37.3\%), frequently misclassifying non-contradictions as contradictions. The LLM-only model is more conservative, yielding higher precision (74.4\%) but slightly lower recall (76.3\%). The hybrid approach outperforms both, achieving\textbf{ 92.0\% accuracy} and an \textbf{F1 score of 89.5\%,} demonstrating that combining NLI-based filtering with LLM judgment reduces false positives while maintaining strong recall.  

Pairwise contradiction detection proves substantially more challenging. The NLI-only model again exhibits high recall (66.7\%) but extremely poor precision (16.0\%), while the LLM-only model improves performance (F1 = 46.9\%) but tends to over-predict contradictions. The hybrid method achieves the best overall balance, reaching \textbf{89.5\% accuracy} and an \textbf{F1 score of 70.9\%}, showing robustness in cross-document settings where subtle contextual shifts and entity references complicate detection. These results underscore that self-contradictions are easier to capture, whereas pairwise contradictions remain significantly harder, even for the hybrid model.

\begin{table}[ht]
\tiny
\centering

\label{tab:contradiction_results}
\begin{tabular}{lcccccccc}
\toprule
 & \multicolumn{4}{c}{\textbf{Self-Contradiction}} & \multicolumn{4}{c}{\textbf{Pairwise Contradiction}} \\
\cmidrule(lr){2-5} \cmidrule(lr){6-9}
\textbf{Model} & \textbf{A } & \textbf{P} & \textbf{R} & \textbf{F1} 
               & \textbf{A} & \textbf{P} & \textbf{R} & \textbf{F1} \\
\midrule
NLI-only   & 41.0 & 37.00 & 81.6 & 51.2   & 36.3 & 16.0 & 66.7 & 25.9 \\
LLM-only   & 81.0 & 74.4 & 76.3 & 75.3   & 66.7 & 31.9 & 88.2 & 46.9 \\
Hybrid     & 92.0 & 89.5 & 89.5 & 89.5   & 89.5 & 66.1 & 76.5 & 70.9 \\
\bottomrule
\end{tabular}
\vspace{+4pt}
\caption{Evaluation results for Contradiction detection (in \%)}
\vspace{-8pt}
\end{table}

Key insights emerge from our findings. NLI models provide high recall but suffer from low precision, whereas LLMs offer better contextual reasoning at the cost of coverage. The hybrid approach integrates their strengths, achieving consistently superior precision and recall. While hybrid or LLM-based methods effectively handle single-document consistency, cross-document contradiction detection remains challenging due to contextual shifts and fragmented evidence. This highlights that we are far from solving cross-document contradiction, underscoring the need for robust, realistic benchmarks like ours that expose subtle, retrieval-induced conflicts in high-stakes domains. Addressing this gap will require retrieval-aware prompting or multi-hop reasoning over evidence.

\vspace{-10pt}
\section{Conclusion and Future Work}
\vspace{-10pt}
This work introduces a framework for systematic \textit{contradiction injection} within long-form legal documents, extending beyond short sentence-pair benchmarks. By generating and labeling contradictions at the document scale, the framework enables rigorous evaluation of both intra-document and cross-document consistency, addressing a critical gap in legal AI. Results from our evaluation of contradiction detection methods show that hybrid NLI+LLM methods achieve the best balance of precision and recall, especially for self-contradictions, whereas cross-document contradictions remain challenging due to contextual variation.

\textbf{Limitations.} One of the main limitations is the high API cost that is proportional to the number of documents we generate and the contradictions we inject. The evaluation relies on \textit{synthetic but realistic} documents rather than large-scale proprietary corpora, limiting validation on naturally occurring contradictions. Human annotation is required for reliable ground-truth labeling, preventing full automation. The focus is on demonstrating controllable contradiction generation rather than exhaustive comparisons of LLM performance.

\textbf{Future Work.} Future directions include: (1) extending the framework into standardized benchmark suites that combine synthetic and real-world contradictions to overcome the current reliance on a limited corpus and enable broader, more representative model evaluation; (2) improving cross-document contradiction detection through retrieval-aware prompting and multi-hop reasoning to align evidence across heterogeneous sources; (3) systematically testing the performance of different LLMs as context validators; and (4) evaluating the ability and robustness of RAG pipelines when presented with conflicting evidence to surface, resolve, or isolate contradictions during generation.

\bibliographystyle{plainnat}
\bibliography{references}

\appendix
\section{Appendix}
\label{appendix}
\subsection{Prompts for Data Generation}
\label{appendix:contentgen}

\begin{tcolorbox}[breakable,colback=white,
                  colframe=blue!50!white,
                  title=Prompt: Base Document Generation,
                  rounded corners,
                  boxrule=0.5mm]

Generate a professional, coherent business document (4--6 paragraphs) for \{company\} in the domain of \{domain\}. 
The topic is `\{topic\}', related to `\{phrase\}'. 
Use the format of a \{doc\_type\}. Structure the document with 4--6 clearly titled sections, paragraphs, numbered lists, or bullets as needed with realistic internal formatting. 
\begin{itemize}
    \item Include at least one \textbf{CLEAR, FACTUAL, ASSERTIVE} sentence early in the document - ideally in paragraph 1 or 2. This sentence will be contradicted later, so make it specific and important. Do not add contradictory information yet.   
    \item The content should create dates and time intervals close to \{date\} and associate them with any requirements. The content should create specific personnel titles and contact info for \{department\}. The content should create specific sub-locations within \{locations\}.
    \item Vary the audience tone between internal staff, legal reviewers, or external partners. Use diverse names, avoiding placeholders like ``John Smith''. Include names appropriate to a global workforce. Be assertive: state policies directly. Avoid vague phrases like ``such as'' or ``for example''. 
    \item If a process is mentioned, describe it in detail (e.g., steps, links, roles, submission method). 
    
\end{itemize}

At the end, include:
\begin{itemize}
    \item \textbf{NEW PEOPLE META DATA:} Names, roles, and affiliations of people mentioned in the document.
    \item \textbf{NEW DOCUMENT META DATA:} Titles and dates of referenced documents (e.g., forms, policy memos, review notices).
\end{itemize}
\end{tcolorbox}

\subsection{Prompts for Contradiction Injection}
\label{appendix:contra}

\begin{tcolorbox}[breakable,colback=white,
                  colframe=blue!50!white,title=Prompt: Identify Statement,rounded corners, boxrule=0.5mm]

You are analyzing a business document to identify the most important/specific statement for contradiction injection. From the following document, select ONE sentence that is:
\begin{itemize}
    \item Clear, factual, and assertive
    \item Important to the document's main topic
    \item Specific (contains numbers, dates, procedures, or concrete details)
\end{itemize}
Return ONLY the selected sentence, nothing else.
\medskip
\textbf{DOCUMENT:}
\textit{\{document\_text\}}
\end{tcolorbox}

\begin{tcolorbox}[breakable,colback=white,
                  colframe=blue!50!white,title=Prompt: Generate Contradiction,rounded corners, boxrule=0.5mm]

You are helping generate business documents that contain realistic internal contradictions. You will be given the original document and one important sentence selected from it (the \textit{target statement}). Your task:
\begin{itemize}[leftmargin=*]
    \item Write 1 or 2 sentences that \textbf{DIRECTLY CONTRADICTS} the target statement after selecting one of the contradiction types mentioned in the few-shot examples.
    \item The contradiction should be \textbf{HARD and DIRECT}, not soft or hedging. Use the same tone and style as the original document.
    \item Do \textbf{NOT} use words like `however', `while', `although', `but', `may', `might', `could', `sometimes', `certain', `extended', `flexibility'.
    \item Make a clear, confident business statement that contradicts the target.
    \item The contradiction should be realistic, specific, and plausible within a formal enterprise setting.
    \item The paragraph should sound like a natural part of the document, not an exception or explanation.
\end{itemize}

\medskip
\textbf{FEW-SHOT EXAMPLES:}

\textit{\{few\_shot\_examples\}} 

\medskip
\textbf{TARGET STATEMENT:}

\textit{\{target\_statement\}}

\medskip
\textbf{ORIGINAL DOCUMENT:}

\textit{\{document\_text\}}
\end{tcolorbox}

\begin{tcolorbox}[breakable,colback=white,
                  colframe=blue!50!white,title=Prompt: Blend Contradiction into Document, rounded corners, boxrule=0.5mm]
You are an expert business writer. You are given a business document and a contradiction paragraph. Your task is to rewrite the document, blending the contradiction paragraph into the most natural place in the document, ideally away from the original target statement. 

Instructions:
\begin{itemize}[leftmargin=*]
    \item Keep the original target statement in the document.
    \item You may rewrite or rephrase the document as needed, but the contradiction must be present and subtle.
    \item The final document should flow naturally, with the contradiction paragraph integrated realistically and not artificially or as an explanation. Use the same tone and style as the original document.
    \item The contradiction must be \textbf{CLEAR and FACTUAL} - do not explain, soften, justify, or hedge.
    \item Do \textbf{NOT} use words like `however', `while', `although', `but', `may', `might', `could', `sometimes', `certain', `extended', `flexibility'.
    \item Do \textbf{NOT} include explanations, reasoning, timeline adjustments, or supporting logic. Just state the conflicting fact.
\end{itemize}

At the end, include:

\begin{itemize}[leftmargin=*]
    \item \textbf{NEW PEOPLE META DATA:} Names, roles, and affiliations of people mentioned in the document.
    \item \textbf{NEW DOCUMENT META DATA:} Titles and dates of referenced documents (e.g., forms, policy memos, review notices).
\end{itemize}

\medskip
\textbf{DOCUMENT:}

\textit{\{base\_content\}}

\medskip
\textbf{ORIGINAL TARGET STATEMENT:}

\textit{\{target\_statement\}}

\medskip
\textbf{CONTRADICTION PARAGRAPH:}

\textit{\{contradiction\_paragraph\}}

\medskip
Return ONLY the final, blended document.
\end{tcolorbox}

\subsection{Prompt used for LLM Contradiction Judge}
\label{appendix:judge}
\begin{tcolorbox}[breakable,colback=white,
                  colframe=blue!50!white,
                  title=Prompt: LLM Contradiction Judge,
                  coltitle=white,
                  rounded corners,
                  boxrule=0.6pt]


You are an expert at detecting logical contradictions in text. Analyze the following two sentences and determine if they contradict each other.

\medskip
\textbf{Sentence 1:} ``\{sentence1\}'' \\
\textbf{Sentence 2:} ``\{sentence2\}''

\medskip
\textbf{Consider:}
\begin{enumerate}
    \item Do these sentences make opposing claims about the same subject?
    \item Are they logically incompatible?
    \item Could both be true in the same context?
    \item Are they discussing different aspects of the same topic?
\end{enumerate}

\medskip
\textbf{Respond with a JSON object containing:}
\begin{itemize}
    \item \texttt{"contradiction"}: true/false
    \item \texttt{"reasoning"}: brief explanation of your decision
    \item \texttt{"confidence"}: float in [0.0--1.0] representing certainty
\end{itemize}

\medskip
\textbf{Example response:}
\begin{verbatim}
{"contradiction": false, 
 "reasoning": "Both sentences discuss 
 different aspects of compliance 
 without conflicting",
 "confidence": 0.8}
\end{verbatim}

\end{tcolorbox}

\end{document}